\documentclass{article}

\usepackage[nonatbib,final]{nips_2016} 

\usepackage[utf8]{inputenc} 
\usepackage[T1]{fontenc}    
\usepackage{hyperref}       
\usepackage{url}            
\usepackage{booktabs}       
\usepackage{amsfonts}       
\usepackage{nicefrac}       
\usepackage{microtype}      
\usepackage{graphicx}
\usepackage{subfig}
\usepackage{adjustbox}
\def\bit{\begin{itemize}}
\def\eit{\end{itemize}}
\def\it{\item}

\def\ben{\begin{enumerate}}
\def\een{\end{enumerate}}

\title{Dense Prediction on Sequences with Time-Dilated Convolutions for Speech Recognition}

\author{
  Tom Sercu \\
  Multimodal Algorithms and Engines Group\\
  IBM T.J Watson Research Center, USA\\
  \texttt{Tom.Sercu1@ibm.com} \\
  \And
  Vaibhava Goel \\
  Multimodal Algorithms and Engines Group\\
  IBM T.J Watson Research Center, USA\\
  \texttt{vgoel@us.ibm.com} \\
}

\begin{document}

\maketitle

\begin{abstract}
In computer vision pixelwise \emph{dense prediction} is the task of predicting a label for each pixel in the image.
Convolutional neural networks achieve good performance on this task, while being computationally efficient.
In this paper we carry these ideas over to the problem of assigning a sequence of labels to a set of speech frames,
a task commonly known as framewise classification.
We show that dense prediction view of framewise classification offers several advantages and insights,
including computational efficiency and the ability to apply batch normalization.
When doing dense prediction we pay specific attention to strided pooling in time and introduce an asymmetric dilated convolution, 
called time-dilated convolution, that allows for efficient and elegant implementation of pooling in time.
We show results using time-dilated convolutions in a very deep VGG-style CNN with batch normalization
on the Hub5 Switchboard-2000 benchmark task.
With a big n-gram language model, we achieve 7.7\% WER which is the best single model single-pass performance reported so far.
\end{abstract}

\section{Introduction}
Deep convolutional networks \cite{lecun1998gradient} have seen tremendous sucess both in 
computer vision \cite{krizhevsky2012imagenet, sermanet2013overfeat, simonyan2014very}
and speech recognition \cite{abdel2012applying, sainath2013deep, sercu2015very}
over the last years.
Many computer vision problems fall into one of two problem types:
the first is classification, where a single label is produced per image,
the second dense pixelwise prediction, where a label is produced for each pixel in the image.
Examples of dense prediciton are semantic segmentation, depth map prediction, optical flow, surface normal prediction, etc.
Efficient convolutional architectures allow to produce a full image sized output rather than predicting
the values for each pixel separately from a small patch centered around the pixel.
In this paper we argue that
\textbf{we should look at acoustic modeling in speech as a dense prediction task} on sequences.
This is in contrast to the the usual viewpoint of ``framewise classification'',
indicating the cross-entropy training stage where a context-window is used as input and
the network predicts only for the center frame.
However, during all other stages, we want the acoustic model to be applied to a sequence, and produce
a sequence of predictions. This is the case during sequence training, test time, or in an end-to-end training setting. 
Similar to convolutional architectures for dense prediction in computer vision,
we focus our efforts on convolutional architectures that process an utterance at once and produce a sequence of labels 
as output, rather than ``splicing'' up the utterance, i.e. labeling each frame independently from a small window around it.

There are four main advantages to convolutional architectures that allow efficient evaluation of full utterance
(without need of splicing) in this dense prediction viewpoint:
\bit
\it Computational efficiency: processing a spliced utterance
requires \verb$window_size$ times more floating point operations.
\it Batch normalization can easily be adopted during sequence training (or end to end training),
which we will show gives strong improvements (as outlined in \cite{sercu2016advances}).
\it The main architectural novelty of this paper is that we can \textbf{allow for strided pooling in time}.
In the next two sections, we will adopt a recent technique from dense prediction, 
named dilated convolutions, for CNN acoustic models to enable strided pooling in time.
Experiments and results for this new model are in section \ref{sec:results}.
\it We will show a unifying viewpoint with Stacked Bottleneck Networks,
and discuss the relevance for end-to-end models with convolutional layers in section \ref{sec:relation}.
\eit

\section{Related work: Pooling in CNNs for dense prediction on images}
Pooling with stride is an essential ingredient of any classification CNN, allowing to
access more context on higher feature maps, while reducing the spatial resolution 
before it is absorbed into the fully connected layers.
However, for dense pixelwise prediction tasks, it is less straightforward how to deal with downsampling:
on the one hand downsampling allows for a ``global view'' by having large receptive fields at low resolution,
on the other hand we also need detail on a small scale, i.e. we need the high resolution information.

To incorporate both global and local information,
downsampling pooling has been incorporated in dense prediction networks in several ways.
Firstly, many methods involve upsampling lower resolution feature maps, usually
combined with some higher resolution feature maps.
In \cite{farabet2012scene}, an image is processed at three different scales
with three different CNNs, after which the output feature maps are merged.
The Fully Convolutional Networks (FCNs) from \cite{long2014fully} use a VGG classification network
as basis, introducing skip connections to merge hi-res lower layers with upsampled low-res
layers from deeper in the network.
SegNet \cite{badrinarayanan2015segnet} uses a encoder-decoder structure,
in which upsampling is done with max-unpooling \cite{zeiler2011adaptive}, 
i.e. by remembering the max location of the encoder's pooling layers.
A second way of using CNNs with strided pooling for dense prediction was proposed in \cite{sermanet2013overfeat}:
at every pooling layer with stride $s \times s$, the input is duplicated $s \times s$ times,
but shifted with offset $(\Delta_x, \Delta_y) \in [0 \dots s-1] \times [0 \dots s-1]$.
After the convolutional stages, the output is then interleaved to recover the full resolution.
A third way (which we will use) is called spatial dilated convolutions, which keeps the feature maps in their original resolution.
The idea is to replace the pooling with stride $s$ by pooling with stride 1,
then dilate all convolutions with a factor $s$, meaning that $\frac{s-1}{s}$ values get skipped.
This was called filter rarefaction in \cite{long2014fully},
introduced as ``d-regularly sparse kernels'' in \cite{li2014highly},
and dubbed spatial dilated convolutions in \cite{yu2015multi}.
It was noted \cite{sermanet2013overfeat, long2014fully} that this method is equivalent to 
shift-and-interleave, though more intuitive.
The recent WaveNet work \cite{oord2016wavenet} uses dilated convolutions for a generative model of audio.

\section{Time-dilated convolutions}
\label{sec:timedilated}
\begin{figure}[b]
    \centering
    \subfloat[Original CNN (XE)]{{
      \includegraphics[height=1.10in]{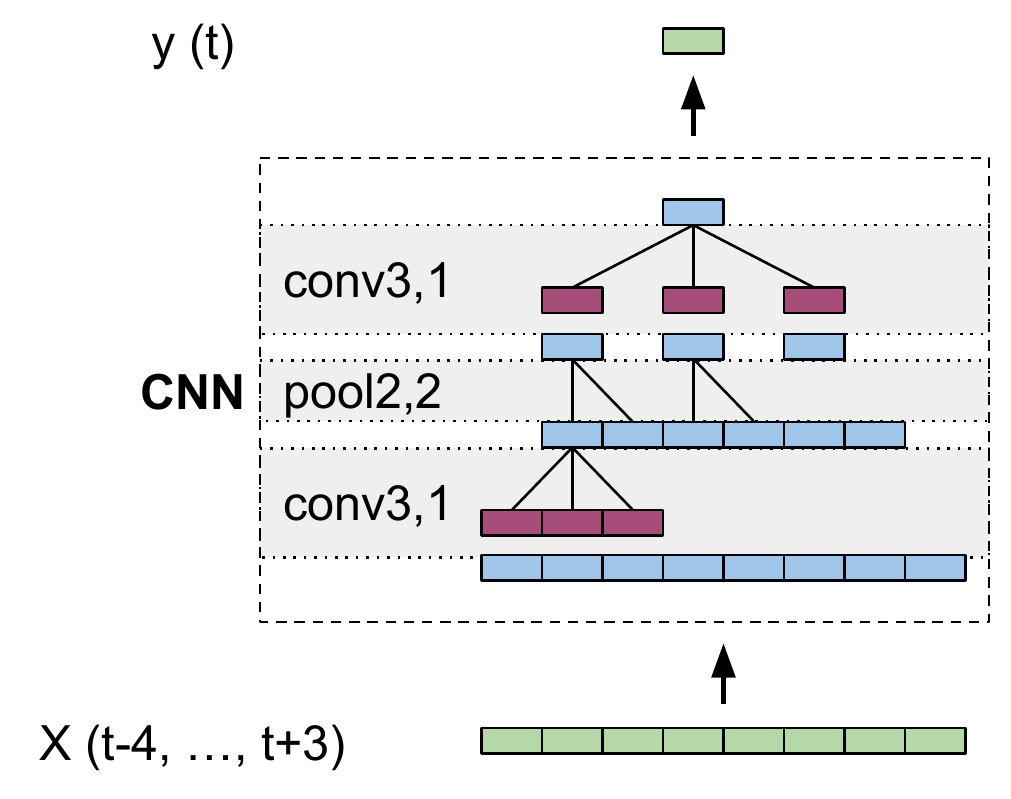}
    }}
    \subfloat[Sequence: Problem]{{
      \includegraphics[height=1.10in]{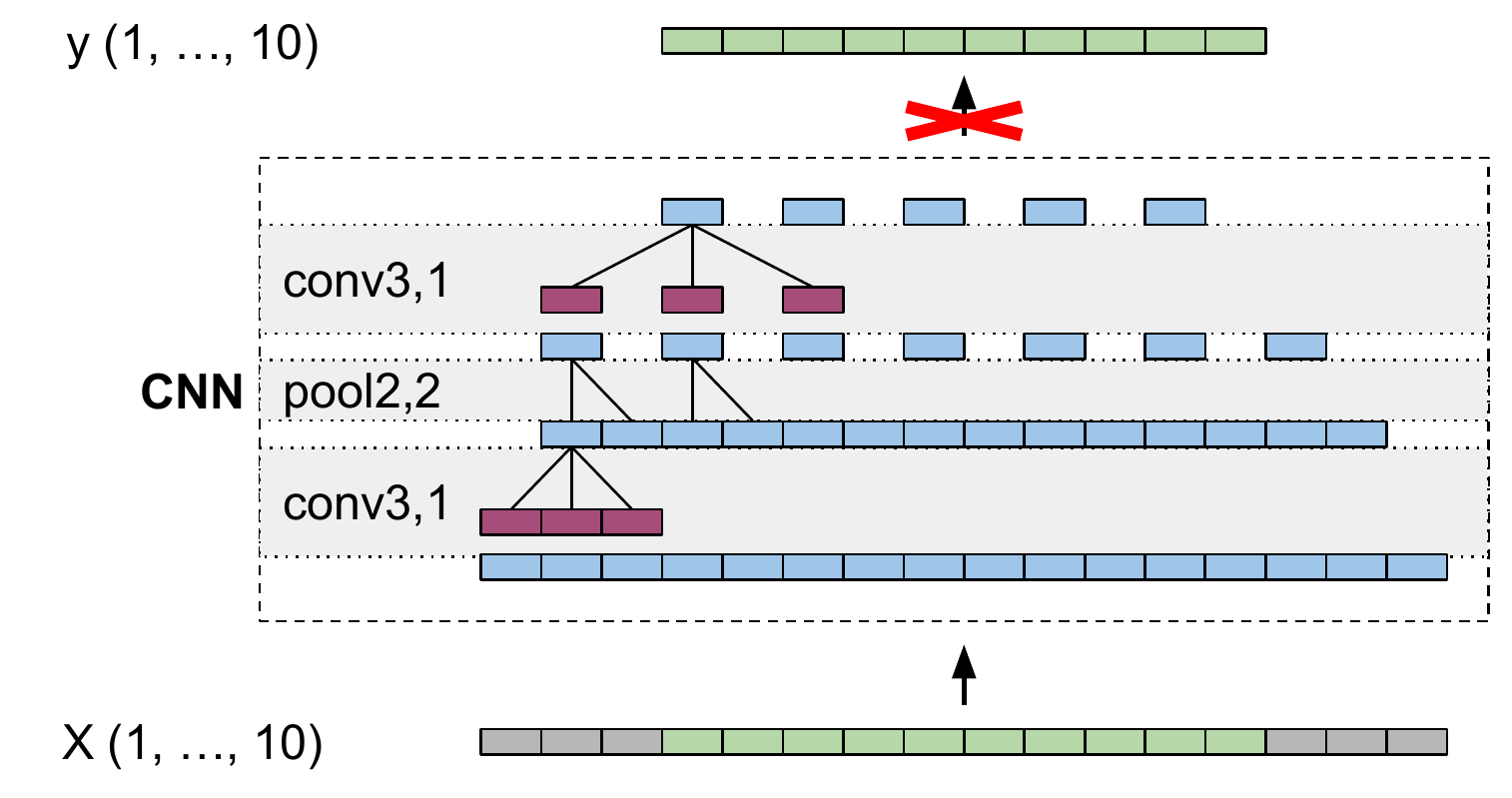}
    }}
    \subfloat[Solution]{{
      \includegraphics[height=1.10in]{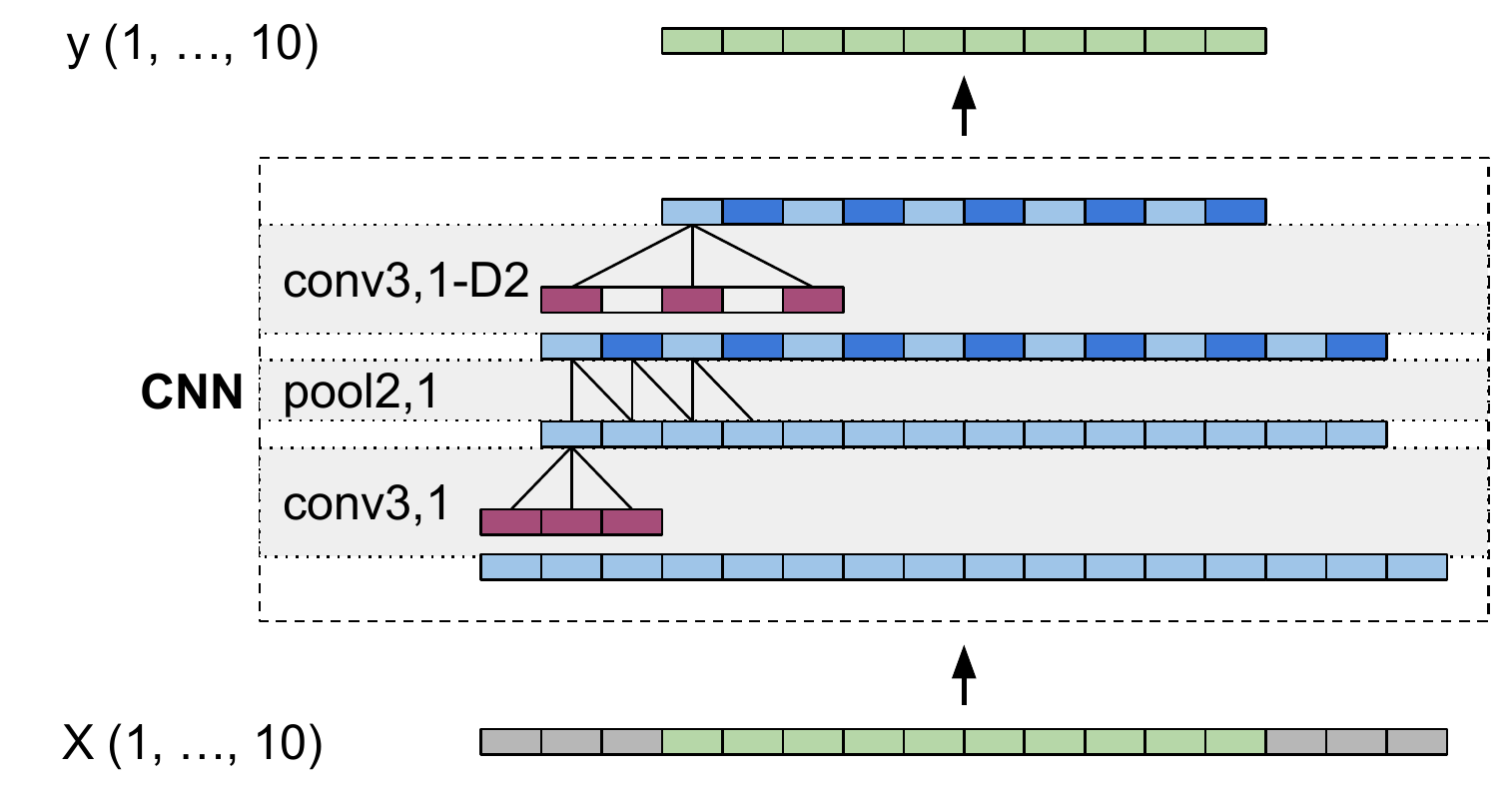}
    }}
    \caption{Example of simple CNN (1 conv, 1 pool, 1 conv layer). 
        Pooling with stride 2 is replaced by pooling with stride 1, while consecutive
        convolutions are dilated with a factor 2.}
    \label{fig:cnn_dilated}
\end{figure}

Previous work on CNNs for acoustic modeling \cite{abdel2012applying, sainath2013deep}
eliminated the possibility of strided pooling in time because of the downsampling effect.
Recent work \cite{sercu2015very, sercu2016advances} shows a significant performance boost 
by using pooling in time during cross-entropy training, however sequence training is prohibitively
expensive since an utterance has to be spliced into \verb$uttLen$ independent windows.
By adapting the notion of dense prediction, we propose to allow pooling in time while maintaining
efficient full-utterance processing, by using an asymmetric version of spatial dilated convolution
with dilation in the time direction but not in the frequency direction,
which we appropriately call time-dilated convolutions.

The problem with strided pooling in time is that the length of the output sequence
is shorter than the length of the input sequence with a factor $2^{(p)}$, assuming $p$
pooling layers with stride 2.
For recurrent end-to-end networks typically a factor 4 size reduction is accepted \cite{amodei2015deep, sainath2015cldnn}
which limits the number of pooling layers to 2, 
while in the hybrid NN/HMM framework, pooling is not acceptable.
Essentially we need a way to do strided pooling in time, while keeping the resolution.
We tackle this problem with a 1D version of sparse kernels \cite{li2014highly}, or equivalently
spatial dilated convolutions \cite{yu2015multi}.

Consider the simple toy CNN (conv3, pool2-s2, conv3) in Figure \ref{fig:cnn_dilated} (a), which takes in a context window of 
8 frames and produces a single output. 
Let's consider applying this CNN to a full utterance of length 10 (padded to length 16), as in figure (b).
The top row of blue outputs is downsampled with factor 2 because of the strided pooling, so the
output sequence length does not match the number of targets (i.e. input size).
The solution of this problem is visualized in Figure \ref{fig:cnn_dilated} (c).
First, we pool without stride, which preserves the resolution after pooling.
However, now our consecutive convolutional layer needs to be modified; 
specifically the kernel has to skip every other value,
in order to ignore the new (dark blue) values which came between the values.
This is dilation (or sparsification) of the kernel with a factor 2 in the time direction.
Formally a 1-D discrete convolution $*_l$ with dilation $l$ which convolves signal $F$ 
with kernel $k$ with size $r$ is defined as 
$(F *_l k)(p) = \sum_{s+lt=p} F(s) k(t) , \quad t \in [-r,r]$.

In general, the procedure to change a CNN with time-pooling from the cross-entropy training (classification)
to dense prediction stage for sequence training and testing is as follows.
Change pooling layers from stride $s$ to stride $1$, and multiply the dilation factor
of all following convolutions with factor $s$.
After this, any convolution coming after $p$ pooling layers with original stride $s$,
will have the dilation factor $s^p$.
Fully connected layers are equivalent to, and can be trivially replaced by, convolutional
layers with kernel $1\times1$ (except the first convolution which has kernel size matching
the output of the conv stack before being flattened for the fully connected layers).
This dilating procedure is how a VGG classification network is adapted for semantic segmentation \cite{li2014highly, yu2015multi}.

Using time-dilated convolutions, the feature maps and output can keep the full resolution of the input,
while pooling with stride.
With pooling, the receptive field in time of the CNN can be larger than the same network without pooling.
This allows to combine the performance gains of pooling \cite{sercu2015very},
while maintaining the computational efficiency and ability to apply batch normalization \cite{sercu2016advances}.

\begin{table*}[b]
  \parbox{0.30\linewidth}{
    \centering
    \begin{adjustbox}{width=0.24\textwidth}
      \small
      \begin{tabular}{| l | p{1.8cm} | }
\hline
  Layer                        & Output: \newline fmaps $\times$ f $\times$ T  \\ \hline
Input window                   &  $  3 \times 64 \times 48$ \\ 
\hline
 conv $7 \times 7$             &  $ 64 \times 64 \times 42$ \\ 
 pool $2 \times 1$             &  $ 64 \times 32 \times 42$ \\ 
\hline
 conv $3 \times 3$             &  $ 64 \times 32 \times 40$ \\ 
 conv $3 \times 3$             &  $ 64 \times 32 \times 38$ \\ 
 conv $3 \times 3$             &  $ 64 \times 32 \times 36$ \\ 
 pool $2 \times 1$             &  $ 64 \times 16 \times 36$ \\ 
\hline
 conv $3 \times 3$             &  $128 \times 16 \times 34$ \\ 
 conv $3 \times 3$             &  $128 \times 16 \times 32$ \\ 
 conv $3 \times 3$             &  $128 \times 16 \times 30$ \\ 
 pool $2 \times 1$             &  $128 \times  8 \times 30$ \\ 
\hline
 conv $3 \times 3$             &  $256 \times  8 \times 28$ \\ 
 conv $3 \times 3$             &  $256 \times  8 \times 26$ \\ 
 conv $3 \times 3$             &  $256 \times  8 \times 24$ \\ 
 pool $2 \times 2$             &  $256 \times  4 \times 12$ \\ 
\hline
 conv $3 \times 3$             &  $512 \times  4 \times 10$ \\ 
 conv $3 \times 3$             &  $512 \times  4 \times  8$ \\ 
 conv $3 \times 3$             &  $512 \times  4 \times  6$ \\ 
 pool $2 \times 2$             &  $512 \times  2 \times  3$ \\ 
\hline
 $3 \times$ FC                 &  2048 \\
 FC                            &  1024 \\
 FC                            &  32000 \\
\hline
\end{tabular}
    \end{adjustbox}
    \caption{\label{tab:cnn} CNN architecture. 
     }
  }
  \qquad
  \parbox{0.70\linewidth}{
    \begin{adjustbox}{width=0.55\textwidth}
      \small
      \begin{tabular}{|l | r | r | r | r |}
  \hline
    & \multicolumn{2}{|c|}{SWB} & \multicolumn{2}{|c|}{CH} \\ \hline
                                              & XE   & ST   & XE   & ST   \\ \hline
    Classic 512 CNN \cite{soltau2014joint}    & 12.6 & 10.4 &      &      \\
    IBM 2016 RNN+VGG+LSTM \cite{saon2016ibm}  &      & 8.6 $^\dagger$ &  & 14.4 $^\dagger$ \\
    MSR 2016 ResNet * \cite{xiong2016microsoft}    &      & 8.9  &      &      \\
    MSR 2016 LACE * \cite{xiong2016microsoft}      &      & 8.6  &      &      \\
    MSR 2016 BLSTM * \cite{xiong2016microsoft}     &      & 8.7  &      &      \\
    \hline
    VGG (pool, inefficient) \cite{saon2016ibm}& 10.2 & 9.4  & 16.3 & 16.0 \\ 
    VGG (no pool) \cite{sercu2016advances}    & 10.8 & 9.7  & 17.1 & 16.7 \\
    VGG-10 + BN (no pool) \cite{sercu2016advances}& 10.8 & 9.5 & 17.0 & 16.3 \\
    \hline
    VGG-13 + BN (no pool)                    & 10.3  & 9.0   & 16.5   & 16.4    \\  
    VGG-13 + BN + pool                       & 9.5  &\bf{8.5}&15.1& 15.4 \\  
    VGG-13 + BN + pool (uncouple CH acwt)    &      &      & \bf{14.8} & 15.2  \\  
    \hline
\end{tabular}

    \end{adjustbox}
    \caption{\label{tab:res} Results with small LM (4M n-grams)}
    \begin{adjustbox}{width=0.42\textwidth}
      \small
      \begin{tabular}{| l | r | r | }
  \hline
                                              & SWB  & CH   \\ \hline
    IBM 2015 DNN+RNN+CNN \cite{saon2015ibm}   & 8.8 $^\dagger$  & 15.3 $^\dagger$ \\
    IBM 2016 RNN+VGG+LSTM \cite{saon2016ibm}  & 7.6 $^\dagger$  & 13.7 $^\dagger$ \\
    MSR 2016 ResNet \cite{xiong2016microsoft}      & 8.6  & 14.8 \\
    MSR 2016 LACE \cite{xiong2016microsoft}        & 8.3  & 14.8 \\
    MSR 2016 BLSTM \cite{xiong2016microsoft}       & 8.7  & 16.2 \\
    \hline
    VGG-13 + BN (no pool)                    & 8.1      & 15.9    \\  
    VGG-13 + BN + pool                       & \bf{7.7} & 14.5 \\  
    VGG-13 + BN + pool (uncouple CH acwt)        &      & \bf{14.4} \\  
    \hline
\end{tabular}

    \end{adjustbox}
    \caption{\label{tab:res2} Results with big LM (36M n-grams)}
  }
\end{table*}

\section{Experiments and results}
\label{sec:results}
We trained a VGG style CNN \cite{simonyan2014very} in the hybrid NN/HMM setting on the 
2000h Switchboard+Fisher dataset.
The architecture and training method is similar to our earlier papers \cite{sercu2015very, sercu2016advances},
and is based on the setup described in \cite{saon2015ibm}.
Our input features are VTLN-warped logmel with $\Delta, \Delta \Delta$, 
the outputs are 32k tied CD states from forced alignment.
Table \ref{tab:cnn} fully specifies the CNN when training on windows and predicting the center frame.
Corresponding to the observations in \cite{sercu2016advances}, we do not pad in time,
though we do pad in the frequency direction.
Training followed the standard two-stage scheme, with first 1600M frames of cross-entropy training (XE) 
followed by 310M frames of Sequence Training (ST).
XE training was done with SGD with nesterov acceleration, with learning rate decaying 
from 0.03 to $9e^{-4}$ over 600M frames.
We use the data balancing from \cite{sercu2015very} with exponent $\gamma=0.8$.
We report results on Hub5'00 (SWB and CH part) after decoding using the standard 
small 4M n-gram language model with a 30.5k word vocabulary.
We saw slight improvement in results when decoding
with exponent $\gamma$ on the prior lower than what is used during training.
As mentioned in section \ref{sec:timedilated}, we use batch normalization in our network,
where the mean and variance statistics are accumulated over both the feature maps and
the frequency direction.
The selection of models, decoding prior and acoustic weight happened by decoding on rt02 as heldout set.

The result after XE and ST are presented in Tables \ref{tab:res} and \ref{tab:res2}.
Baseline with * from personal communication with the authors.
Baseline with $\dagger$ means system combination.
Note that the baselines from \cite{xiong2016microsoft} use slightly smaller LMs:
3.4M n-grams for small LM (table \ref{tab:res}) and 16M n-grams for big LM (table \ref{tab:res2}).
We note that one typically does subsequent rescoring with more advanced language models like RNN or LSTM LMs;
this way in \cite{xiong2016achieving} a single model performance of 6.6 is achieved, starting from 8.6.
With just n-gram decoding, this result is to our knowledge the best published single model.

\section{Relation to other models}
\label{sec:relation}
\textbf{Stacked bottleneck networks} (SBN) \cite{grezl2009investigation, vesely2011convolutive, grezl2014adaptation}
or hierarchical bottleneck networks \cite{plahl2010hierarchical}
are a influential acoustic model in hybrid NN/HMM speech recognition.
SBNs are typically seen as two consecutive DNNs, each stage separately trained discriminatively with
a bottleneck (small hidden layer). The first DNN sees the input features, while the second
DNN gets the bottleneck features from the first DNN as input.
Typically, the second DNN gets 5 bottleneck features with stride 5, i.e. features from position
$\{-10, -5, 0, 5, 10\}$ relative to the center \cite{grezl2014adaptation}.
In \cite{vesely2011convolutive}, it was pointed out that this SBN is convolutional and one can 
backpropagate through both stages together.

In fact this multi-stage SBN architecture is a special case of a CNN with time-dilated convolution.
Specifically, the DNN is equivalent to a CNN with a large first kernel followed by all $1\times1$ kernels.
The second DNN is exactly equivalent to a CNN with the first kernel having size 5 and 
dilation factor 5 in the time direction. 
The layers after the bottleneck in the first DNN form an auxilary classifier.
This realization prompts a number of directions in which the SBNs can be extended.
Firstly, by avoiding the large kernel in the first convolutional layer, it is possible to keep
time and frequency structure in the internal representations in future layers, enabling increased depth.
Secondly, rather than increasing the time-dilation factor to 5 at once, it seems more natural to
gradually increase the time-dilation factor throughout the depth of the network.

Convolutional networks are also used in \textbf{end-to-end models} for speech recognition.
Both the CLDNN architecture \cite{sainath2015cldnn} and Deep Speech 2 (DS2) \cite{amodei2015deep} 
combine a convolutional network as first stage with LSTM and fully connected (DNN) output layers.
In Wav2Letter \cite{collobert2016wav2letter}, a competitive end-to-end model is presented which is fully convolutional.
Both DS2 and Wav2Letter do a certain amount of downsampling through pooling or striding,
which can be accepted when training with a CTC (or AutoSeg \cite{collobert2016wav2letter}) 
criterion since it doesn't require the output to be the same length as the input.
However, DS2 does report a degradation on English, which they work around using grapheme bigram targets.

The time-dilated convolutions we introduced, could improve these end to end models in two ways:
either, one could allow the same amount of pooling while keeping
a higher resolution.
Alternatively, one could keep the same resolution, but expand the receptive field by
adding more time-dilated convolution layers, which gives access to a broader context in the CNN layers.
In conclusion, this work is both relevant to end-to-end models and to hybrid HMM/NN models.

\section{Conclusion}
We drew the parallel between dense prediction in computer vision and
framewise sequence labeling, both in the HMM/NN and end-to-end setting.
This provided us with the tool (time-dilated convolutions) to adopt
pooling in time to CNN acoustic models, while maintaining efficient processing and batch normalization
on full utterances.
On Hub5'00 we brought down the WER from 9.4\% in previous work to 8.5\%, a 10\% relative improvement.
With a big (36M n-gram) language model, we achieve 7.7\% WER, the best single model single-pass performance reported so far.

\newpage
\bibliographystyle{IEEEbib}
{ 
  \fontsize{8pt}{9pt} \selectfont
\bibliography{refs}

\begin{thebibliography}{10}

\bibitem{lecun1998gradient}
Yann LeCun, L{\'e}on Bottou, Yoshua Bengio, and Patrick Haffner,
\newblock ``Gradient-based learning applied to document recognition,''
\newblock {\em Proceedings of the IEEE}, vol. 86, no. 11, pp. 2278--2324, 1998.

\bibitem{krizhevsky2012imagenet}
Alex Krizhevsky, Ilya Sutskever, and Geoffrey~E Hinton,
\newblock ``Imagenet classification with deep convolutional neural networks,''
\newblock in {\em Advances in neural information processing systems}, 2012, pp.
  1097--1105.

\bibitem{sermanet2013overfeat}
Pierre Sermanet, David Eigen, Xiang Zhang, Micha{\"e}l Mathieu, Rob Fergus, and
  Yann LeCun,
\newblock ``Overfeat: Integrated recognition, localization and detection using
  convolutional networks,''
\newblock {\em arXiv:1312.6229}, 2013.

\bibitem{simonyan2014very}
Karen Simonyan and Andrew Zisserman,
\newblock ``Very deep convolutional networks for large-scale image
  recognition,''
\newblock {\em CoRR arXiv:1409.1556}, 2014.

\bibitem{abdel2012applying}
Ossama Abdel-Hamid, Abdel-rahman Mohamed, Hui Jiang, and Gerald Penn,
\newblock ``Applying convolutional neural networks concepts to hybrid nn-hmm
  model for speech recognition,''
\newblock in {\em Proc. ICASSP}, 2012.

\bibitem{sainath2013deep}
Tara~N Sainath, Abdel-rahman Mohamed, Brian Kingsbury, and Bhuvana Ramabhadran,
\newblock ``Deep convolutional neural networks for lvcsr,''
\newblock in {\em Proc. ICASSP}, 2013.

\bibitem{sercu2015very}
Tom Sercu, Christian Puhrsch, Brian Kingsbury, and Yann LeCun,
\newblock ``Very deep multilingual convolutional neural networks for lvcsr,''
\newblock {\em Proc. ICASSP}, 2016.

\bibitem{sercu2016advances}
Tom Sercu and Vaibhava Goel,
\newblock ``Advances in very deep convolutional neural networks for lvcsr,''
\newblock {\em Proc. Interspeech}, 2016.

\bibitem{farabet2012scene}
Cl{\'e}ment Farabet, Camille Couprie, Laurent Najman, and Yann LeCun,
\newblock ``Scene parsing with multiscale feature learning, purity trees, and
  optimal covers,''
\newblock {\em Proc. ICML}, 2012.

\bibitem{long2014fully}
Jonathan Long, Evan Shelhamer, and Trevor Darrell,
\newblock ``Fully convolutional networks for semantic segmentation,''
\newblock {\em CVPR}, 2015.

\bibitem{badrinarayanan2015segnet}
Vijay Badrinarayanan, Alex Kendall, and Roberto Cipolla,
\newblock ``Segnet: A deep convolutional encoder-decoder architecture for image
  segmentation,''
\newblock {\em arXiv:1511.00561}, 2015.

\bibitem{zeiler2011adaptive}
Matthew~D Zeiler, Graham~W Taylor, and Rob Fergus,
\newblock ``Adaptive deconvolutional networks for mid and high level feature
  learning,''
\newblock in {\em 2011 International Conference on Computer Vision}. IEEE,
  2011, pp. 2018--2025.

\bibitem{li2014highly}
Hongsheng Li, Rui Zhao, and Xiaogang Wang,
\newblock ``Highly efficient forward and backward propagation of convolutional
  neural networks for pixelwise classification,''
\newblock {\em arXiv:1412.4526}, 2014.

\bibitem{yu2015multi}
Fisher Yu and Vladlen Koltun,
\newblock ``Multi-scale context aggregation by dilated convolutions,''
\newblock {\em proc ICLR}, 2016.

\bibitem{oord2016wavenet}
Aaron van~den Oord, Sander Dieleman, Heiga Zen, Karen Simonyan, Oriol Vinyals,
  Alex Graves, Nal Kalchbrenner, Andrew Senior, and Koray Kavukcuoglu,
\newblock ``Wavenet: A generative model for raw audio,''
\newblock {\em arXiv:1609.03499}, 2016.

\bibitem{amodei2015deep}
Dario Amodei, Rishita Anubhai, Eric Battenberg, Carl Case, Jared Casper, Bryan
  Catanzaro, Jingdong Chen, Mike Chrzanowski, Adam Coates, Greg Diamos, et~al.,
\newblock ``Deep speech 2: End-to-end speech recognition in english and
  mandarin,''
\newblock {\em CoRR arXiv:1512.02595}, 2015.

\bibitem{sainath2015cldnn}
Tara~N Sainath, Oriol Vinyals, Andrew Senior, and Ha{\c{s}}im Sak,
\newblock ``Convolutional, long short-term memory, fully connected deep neural
  networks,''
\newblock in {\em proc. ICASSP}, 2015.

\bibitem{soltau2014joint}
Hagen Soltau, George Saon, and Tara~N Sainath,
\newblock ``Joint training of convolutional and non-convolutional neural
  networks,''
\newblock {\em to Proc. ICASSP}, 2014.

\bibitem{saon2016ibm}
George Saon, Tom Sercu, Steven Rennie, and Hong-Kwang~J Kuo,
\newblock ``The ibm 2016 english conversational telephone speech recognition
  system,''
\newblock {\em Proc. Interspeech}, 2016.

\bibitem{xiong2016microsoft}
W~Xiong, J~Droppo, X~Huang, F~Seide, M~Seltzer, A~Stolcke, D~Yu, and G~Zweig,
\newblock ``The microsoft 2016 conversational speech recognition system,''
\newblock {\em arXiv:1609.03528}, 2016.

\bibitem{saon2015ibm}
George Saon, Hong-Kwang~J Kuo, Steven Rennie, and Michael Picheny,
\newblock ``The ibm 2015 english conversational telephone speech recognition
  system,''
\newblock {\em Proc. Interspeech}, 2015.

\bibitem{xiong2016achieving}
W~Xiong, J~Droppo, X~Huang, F~Seide, M~Seltzer, A~Stolcke, D~Yu, and G~Zweig,
\newblock ``Achieving human parity in conversational speech recognition,''
\newblock {\em arXiv:1610.05256}, 2016.

\bibitem{grezl2009investigation}
Frantisek Grezl, Martin Karafi{\'a}t, and Lukas Burget,
\newblock ``Investigation into bottle-neck features for meeting speech
  recognition.,''
\newblock in {\em Proc. Interspeech}, 2009.

\bibitem{vesely2011convolutive}
Karel Vesel{\`y}, Martin Karafi{\'a}t, and Franti{\v{s}}ek Gr{\'e}zl,
\newblock ``Convolutive bottleneck network features for lvcsr,''
\newblock in {\em ASRU}, 2011.

\bibitem{grezl2014adaptation}
Frantisek Gr{\'e}zl, Martin Karafi{\'a}t, and Karel Vesel{\`y},
\newblock ``Adaptation of multilingual stacked bottle-neck neural network
  structure for new language,''
\newblock in {\em Proc. ICASSP}, 2014.

\bibitem{plahl2010hierarchical}
Christian Plahl, Ralf Schl{\"u}ter, and Hermann Ney,
\newblock ``Hierarchical bottle neck features for lvcsr.,''
\newblock in {\em Interspeech}, 2010, pp. 1197--1200.

\bibitem{collobert2016wav2letter}
Ronan Collobert, Christian Puhrsch, and Gabriel Synnaeve,
\newblock ``Wav2letter: an end-to-end convnet-based speech recognition
  system,''
\newblock {\em arXiv:1609.03193}, 2016.

\end{thebibliography}
}

\end{document}